\documentclass[11pt]{article}

\usepackage[final]{acl}

\usepackage{times}
\usepackage{latexsym}

\usepackage[T1]{fontenc}
\usepackage{enumitem}
\usepackage[utf8]{inputenc}

\usepackage{placeins}
\usepackage{microtype}
\usepackage[most]{tcolorbox}
\usepackage{xcolor}
\usepackage{booktabs}
\usepackage{multirow}
\usepackage[table]{xcolor}
\definecolor{DARSGray}{gray}{0.94}
\definecolor{GainGreen}{RGB}{0,120,80}
\newcommand{\gain}[1]{\textcolor{GainGreen}{#1}}
\usepackage{amssymb}

\definecolor{obsbg}{HTML}{F7FAFC}
\definecolor{obsframe}{HTML}{B8C7D9}

\newtcolorbox{observationbox}{
    colback=obsbg,
    colframe=obsframe,
    boxrule=0.45pt,
    arc=1.5pt,
    left=5pt,
    right=5pt,
    top=4pt,
    bottom=4pt,
    enhanced
}
\usepackage{inconsolata}
\usepackage{subcaption}
\usepackage{graphicx}
\usepackage{marvosym}

\title{From Sampled Outcomes to Capability Distributions: \\Rethinking Supervision for LLM Routing}

\author{
Guannan Lai$^{1,2,3}$,
Haoran Hu$^{1,2}$,
Long Chen$^{4}$,
Zhenguo Li$^{4,5}$,
Han-Jia Ye$^{1,2}$\textsuperscript{(\Letter)}
\\
$^{1}$ School of Artificial Intelligence, Nanjing University \\
$^{2}$ National Key Laboratory for Novel Software Technology, Nanjing University \\
$^{3}$ SinapisAI \\
$^{4}$ Hong Kong University of Science and Technology \\
$^{5}$ Frontier Robotics \\
{\tt\small
laign@lamda.nju.edu.cn,
huhr@smail.nju.edu.cn,
longchen@ust.hk,
zhenguol@gmail.com,
yehj@lamda.nju.edu.cn
}
\\
}

\begin{document}

\maketitle

\begin{abstract}
Existing LLM routing methods often construct supervision from a single
sampled response for each query--model pair. Because LLM generation is
stochastic, however, such an observation can be an unstable estimate of
model capability: semantically equivalent query formulations and repeated
decoding may yield different scores and even different model preferences.
We show that this instability can further propagate from routing labels to
learned routing policies.
To address this issue, we propose \textbf{DARS}
(Distribution-Aware Routing Supervision), which estimates query-level
model capability from repeated observations spanning semantics-preserving
query rewrites and stochastic decoding. DARS summarizes expected quality,
expected cost, and performance variability to construct risk-aware
supervision without changing the downstream router architecture.
Experiments across diverse tasks and routing methods show that DARS
generally improves routing utility and cost--quality trade-offs over
single-shot supervision. Further analyses show that its benefits persist
under moderate sampling budgets and different decoding temperatures.
These results suggest that reliable LLM routing should move beyond
individual sampled outcomes and instead model query-level capability
distributions.
\end{abstract}

\section{Introduction}

\begin{figure}[t]
	\centering
	\includegraphics[width=0.9\linewidth]{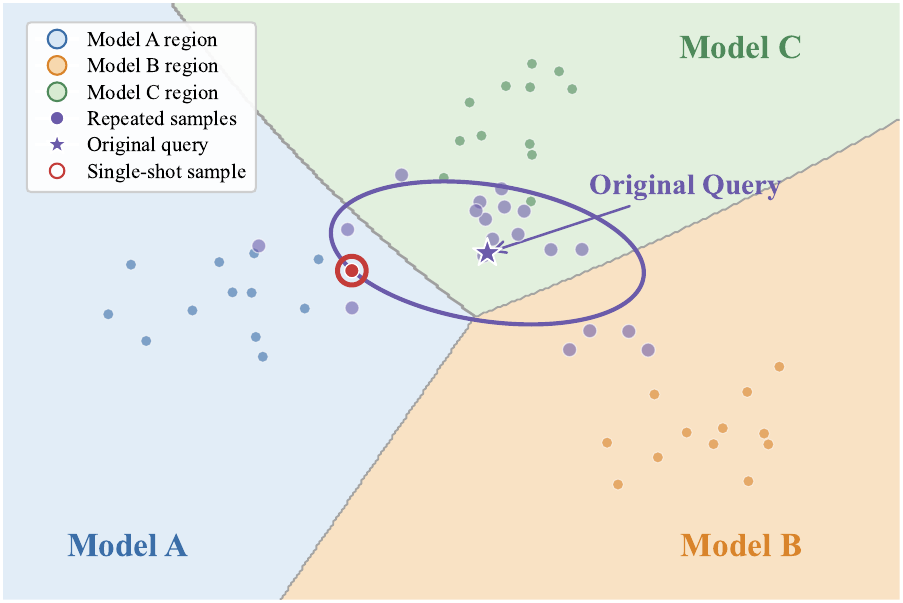}
    \caption{
    Illustration of the single-shot label issue in LLM routing. 
    The colored regions represent different models that are preferred in different parts of the query-behavior space. 
    For the same original query, repeated generations may lead to multiple observed outcomes across these regions, while single-shot supervision only observes one sampled outcome and may therefore produce an unstable routing label.
    }
    \label{fig:intro}
\end{figure}

Large language models (LLMs) are increasingly deployed as heterogeneous model pools rather than isolated single systems. In practice, different models exhibit distinct strengths: lightweight models can often handle simple instructions or factual queries, while larger or more specialized models may be required for complex reasoning, mathematical problem solving, biomedical question answering, or long-context understanding. These models also differ substantially in inference cost, latency, context length, and availability. As a result, \emph{LLM routing} has become an important mechanism for cost-effective deployment: given an input query, a router selects an appropriate model from a candidate pool, aiming to balance response quality and serving cost. Recent routing methods and benchmarks have made substantial progress in learning such query-dependent model selection policies~\citep{chen2024frugalgpt,ong2024routellm,hu2024routerbench,huang2025routereval}.

Despite this progress, most routing supervision still represents each query-model pair with a single generated response and its score. This point-estimate view is convenient, but it is poorly aligned with
free-form LLM generation, where outputs are inherently stochastic. Prior work on uncertainty estimation and hallucination detection has shown that LLM generations can be sensitive to sampling randomness and may vary in both surface form and semantic content~\citep{kuhn2023semantic,manakul2023selfcheckgpt,farquhar2024detecting}. As illustrated in Figure~\ref{fig:intro}, even for the same original query, different sampled outcomes may suggest different model preferences. This raises a fundamental question for LLM routing: can a single sampled response reliably serve as the supervision signal for model selection?

This mismatch can propagate through the entire routing pipeline. Variability in a model's sampled response may alter its observed score; changes in observed scores may change which model appears preferable for a query; and routers trained on such sample-dependent labels may learn policies that reflect incidental generation noise rather than stable differences in model capability. These issues motivate a distribution-aware view of routing supervision, where labels are constructed from repeated observations of query-model behavior rather than from isolated sampled outputs.

In this paper, we propose \textbf{DARS} (\textbf{D}istribution-\textbf{A}ware \textbf{R}outing \textbf{S}upervision), a framework for constructing distribution-aware supervision signals for LLM routing.\footnote{
\href{https://github.com/AIGNLAI/DARS}{Source code} and
\href{https://huggingface.co/datasets/AIGNLAI/DARS}{datasets}
are publicly available.
} DARS considers uncertainty from both the input side and the output side: it uses semantically preserving prompt rewrites to capture sensitivity to query formulation, and repeated decoding to capture stochastic variation in generation. Based on these observations, DARS constructs more reliable routing labels that reflect query-level model behavior rather than isolated outputs. We evaluate DARS on diverse tasks spanning multiple-choice scientific reasoning, mathematical problem solving, and reading comprehension, using a heterogeneous pool of six LLMs. Our analysis shows that single-shot labels can be unstable and misleading for model selection, while a moderate number of repeated observations already provides
substantially more stable supervision. Further experiments show that routers trained with DARS supervision achieve more stable and effective routing behavior than those trained from single-response labels.
Our contributions are threefold:
\begin{itemize}[leftmargin=1.2em,itemsep=0.25em,topsep=0.25em]
    \item We identify and systematically analyze the \emph{single-shot label assumption} in LLM routing, showing that representing query-model behavior with one generated response can introduce unstable and noisy routing supervision.

    \item We introduce \textbf{DARS}, a distribution-aware routing supervision framework that accounts for both input-side and output-side uncertainty through prompt rewrites and repeated decoding.

    \item We provide empirical evidence across diverse datasets and models that distribution-aware supervision yields more reliable routing labels and improves learned routing policies compared with single-response supervision.
\end{itemize}
\section{Related Work}
\label{sec:related_work}

\paragraph{LLM routing.}
LLM routing aims to select an appropriate model from a heterogeneous model pool for each input query, usually balancing response quality and inference cost \cite{lai2026routejudge}. Early work studies cost-aware model selection and cascading strategies, showing that using multiple LLMs adaptively can reduce inference cost while maintaining or even improving performance \citep{chen2024frugalgpt, aggarwal2024automix, yue2024large,lai2026routing}. Recent methods further learn query-dependent routing policies with different supervision and modeling strategies. For example, Hybrid LLM and BEST-Route study adaptive routing under quality-cost constraints \citep{ding2024hybrid, ding2025bestroute}; RouteLLM learns routing from preference data \citep{ong2024routellm}; TensorOpera Router, EmbedLLM, IRT-Router, GraphRouter, and causal routing develop different learned routers based on regression, representation learning, item response theory, graph modeling, or regret minimization \citep{stripelis2024tensoropera, zhuang2025embedllm, song2025irt, feng2025graphrouter, tsiourvas2025causal,ma2026mmr}. Other studies investigate training-free, retrieval-based, controllable, or preference-aware routing strategies \citep{mei2025omnirouter, piskala2025dynamic}. Benchmarks such as RouterBench and RouterEval provide systematic evaluation settings for multi-LLM routing \citep{hu2024routerbench, huang2025routereval}. Unlike these works, we do not focus on designing a new router architecture; instead, we revisit the supervision used to train or evaluate routers.

\paragraph{Uncertainty-aware and efficient LLM inference.}
Several lines of work improve LLM inference efficiency by estimating uncertainty, deciding when to defer to stronger models, or allocating computation adaptively. For example, uncertainty-based two-tier selection and related caching or distillation strategies reduce expensive model calls by identifying when a cheaper model is sufficient \citep{ramrez2024optimising, ramirez2024cache}. Other work studies budget- and quality-controllable routing, quality-of-service-aware routing, or dynamic routing under deployment constraints \citep{mei2025omnirouter, yang2025quality}. Recent studies also examine routing robustness and vulnerabilities, showing that routing policies can be fragile under distribution shifts, adversarial settings, or lifecycle changes \citep{shafran2025rerouting, lin2025life, kassem2026robust}. These works highlight the importance of reliable model selection, but they typically assume that the observed score of a query-model pair is a suitable supervision signal. In contrast, DARS argues that such observations are themselves stochastic and should be treated as samples from an underlying capability distribution.

\paragraph{Uncertainty and robustness in language model generation.}
Our work is also related to uncertainty estimation and robustness analysis for LLM generation. Prior studies show that LLM outputs can vary substantially under sampling randomness, and that inconsistency across generations can reveal uncertainty or hallucination risk \citep{kuhn2023semantic, manakul2023selfcheckgpt, farquhar2024detecting}. Another line of work studies prompt robustness, showing that semantically similar or adversarially perturbed prompts may lead to different model predictions \citep{gan2023sensitivity, zhu2023promptrobust}. These findings suggest that a model's behavior on a query cannot be fully characterized by a single generated response. DARS brings this insight to LLM routing: instead of constructing routing labels from isolated sampled outcomes, it uses query rewrites and repeated stochastic decoding to estimate distribution-aware capability signals for each query-model pair.

\section{Preliminaries}
\label{sec:preliminaries}

\subsection{Single-shot Supervision for LLM Routing}
\label{sec:single_point_supervision}

LLM routing considers a candidate model pool $\mathcal{M}=\{m_1,\ldots,m_K\}$ and aims to select an appropriate model for each input query $x$. Given a routing function $r(\cdot)$, the selected model is $r(x)\in\mathcal{M}$. The goal is to achieve a favorable trade-off between response quality and inference cost: the router should assign difficult queries to more capable models when necessary, while avoiding unnecessary use of expensive models for queries that can be handled by cheaper alternatives.

In a typical routing setup, each model $m$ produces a response $y$ for query $x$. The response is evaluated along two dimensions. First, a task-specific performance function $s(\cdot)$ measures the quality of the response, such as accuracy for multiple-choice or mathematical tasks, or F1 for reading-comprehension tasks. Second, a cost function $\kappa(\cdot)$ measures the inference cost of producing the response, which may depend on the model, the query, the output length, and the decoding process. Thus, for a query-model pair $(x_i,m)$, a generated response induces both a performance observation and a cost observation.

Most existing routing datasets construct supervision from a single observed response for each query-model pair. Specifically, for each query $x_i$ and model $m$, one response is generated:
\[
y_{i,m}^{\mathrm{single}} \sim P(y\mid x_i,m),
\]
and the corresponding single-shot observations are recorded as
\[
q_{i,m}^{\mathrm{single}}
=
s(y_{i,m}^{\mathrm{single}}, x_i),
\quad
c_{i,m}^{\mathrm{single}}
=
\kappa(y_{i,m}^{\mathrm{single}}, x_i, m).
\]
Here, $q_{i,m}^{\mathrm{single}}$ denotes the observed performance of model $m$ on query $x_i$, while $c_{i,m}^{\mathrm{single}}$ denotes the observed cost of this particular generation. Importantly, the cost is not necessarily a model-level constant: even for the same model, it may vary across queries and sampled outputs due to differences in prompt length, completion length, and generation behavior.

Given a dataset $\mathcal{D}=\{x_i\}_{i=1}^{N}$ and a cost budget $C_{\max}$, the routing objective can be formulated as a constrained optimization problem:

\[
\max_{r}
\frac{1}{N}
\sum_{i=1}^{N}
q_{i,r(x_i)}^{\mathrm{single}}
\quad
\mathrm{s.t.}
\quad
\frac{1}{N}
\sum_{i=1}^{N}
c_{i,r(x_i)}^{\mathrm{single}}
\leq C_{\max}.
\]
That is, the router aims to maximize average performance while keeping the average inference cost within a prescribed budget.

This protocol is simple and scalable, which explains its prevalence in routing benchmarks and learned routing systems. However, it implicitly treats one sampled generation as a reliable estimate of both the performance and cost behavior of a query-model pair. The next section examines why this single-shot view can be problematic for LLM routing.

\subsection{Uncertainty in Query-Model Behavior}
\label{sec:uncertainty_sources}

The reliability of single-shot routing supervision depends on whether a query-model pair has stable behavior under natural variations. In this work, we distinguish two sources of uncertainty that are particularly relevant to LLM routing.

\paragraph{Input-side uncertainty.}
The first source comes from variations in how the same query is expressed. In real applications, semantically equivalent user requests may differ in wording, structure, level of detail, or prompt format. Ideally, a model's capability on a query should be stable under such meaning-preserving variations. However, prior work on prompt robustness has shown that LLMs can be sensitive to prompt formulations, and that small changes in prompts may lead to different predictions or performance degradation~\citep{zhu2023promptrobust,gan2023sensitivity}. For routing, this means that the estimated suitability of a model may depend not only on the underlying task, but also on the particular surface form of the input. We refer to this source of variability as \emph{input-side uncertainty}.

\paragraph{Output-side uncertainty.}
The second source comes from stochastic generation itself. Even when the input query and model are fixed, decoding can produce different responses, which may vary in factuality, reasoning path, final answer, or semantic content. Prior work on uncertainty estimation and hallucination detection has shown that such variation is informative: inconsistent generations often indicate higher uncertainty or lower reliability~\citep{kuhn2023semantic,manakul2023selfcheckgpt,farquhar2024detecting}. For routing, this implies that a single generated response may be insufficient to characterize how well a model can handle a query. We refer to this source of variability as \emph{output-side uncertainty}.

\begin{figure*}[t]
    \centering
    \begin{subfigure}[t]{0.32\linewidth}
        \centering
        \includegraphics[width=\linewidth]{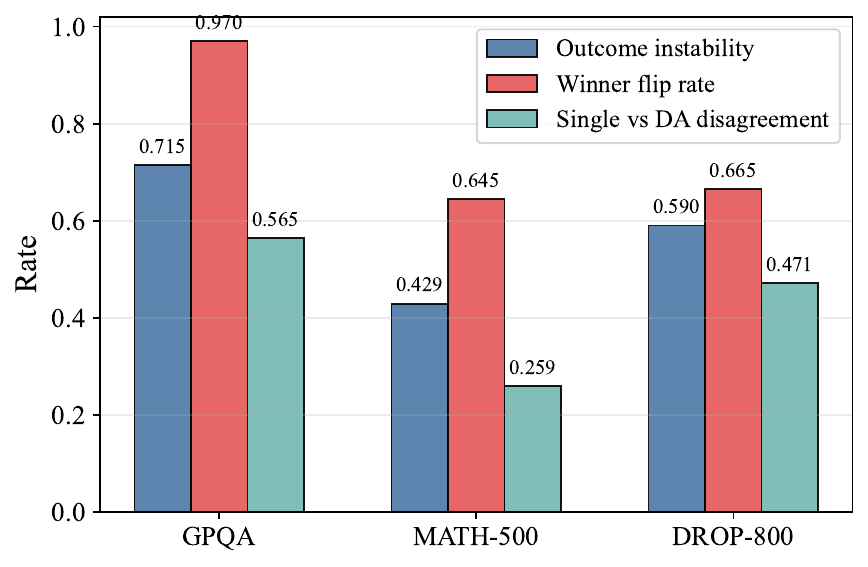}
        \caption{Label instability}
        \label{fig:ana_instability}
    \end{subfigure}
    \hfill
    \begin{subfigure}[t]{0.32\linewidth}
        \centering
        \includegraphics[width=\linewidth]{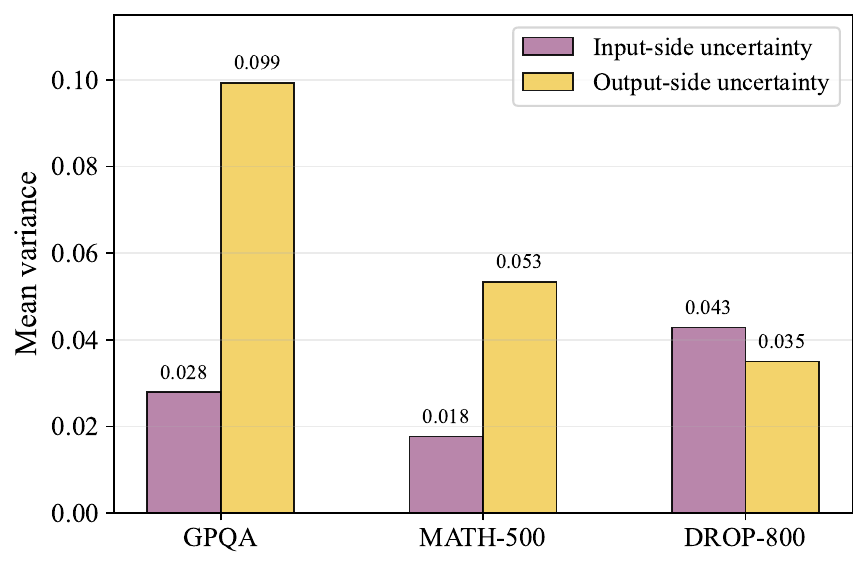}
        \caption{Uncertainty sources}
        \label{fig:ana_uncertainty}
    \end{subfigure}
    \hfill
    \begin{subfigure}[t]{0.32\linewidth}
        \centering
        \includegraphics[width=\linewidth]{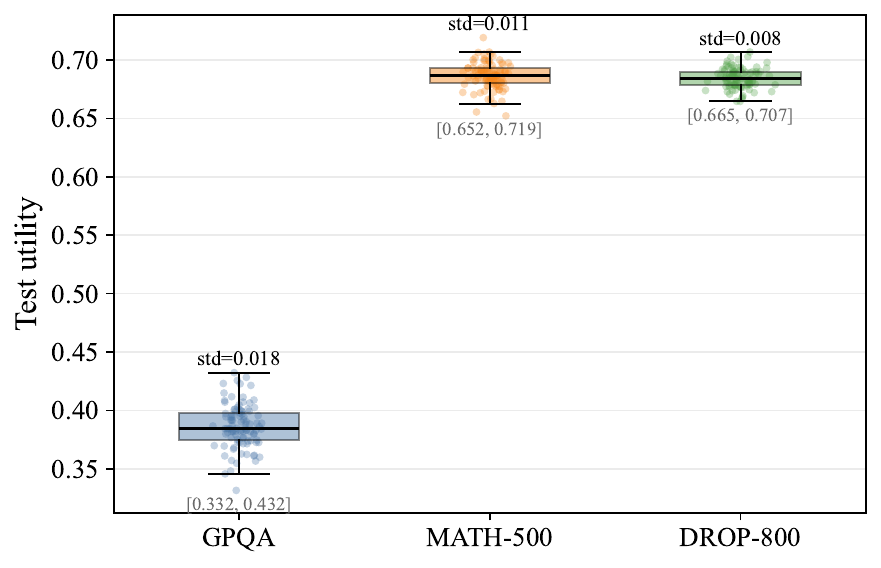}
        \caption{Router variance}
        \label{fig:ana_router_variance}
    \end{subfigure}
    \caption{
    Diagnostic analysis of single-shot routing supervision.
    (a) Single-shot labels are unstable at the outcome, winner, and label levels.
    (b) Input-side and output-side uncertainty contribute differently across datasets.
    (c) Routers trained from different single-shot samples exhibit non-negligible performance variation.
    }
    \label{fig:single_point_limitations}
\end{figure*}

\subsection{Limitations of Single-shot Routing Supervision}
\label{sec:limitations_single_point}

To examine whether single-shot supervision provides reliable routing signals, we conduct a diagnostic analysis across tasks with different output formats and evaluation protocols. This section focuses only on the limitations of existing supervision protocols, leaving the design of distribution-aware routing to the next section.

\paragraph{Experimental setup.}
We use three datasets covering complementary task types. \textbf{GPQA} is a graduate-level multiple-choice question answering benchmark covering biology, physics, and chemistry~\citep{rein2024gpqa}. \textbf{MATH-500} is a 500-problem subset of the MATH benchmark for mathematical problem solving~\citep{hendrycks2021measuring,lightman2024let}. \textbf{DROP-800} is an 800-problem subset of the reading-comprehension benchmark requiring discrete reasoning over paragraphs, such as counting, sorting, and arithmetic reasoning~\citep{dua2019drop}. For each dataset, we sample 200 training queries for constructing repeated observations. To capture input-side uncertainty, we generate five semantically preserving rewrites for each query using GPT-4o~\citep{hurst2024gpt}, while keeping the gold answer and candidate choices unchanged. To capture output-side uncertainty, for each rewritten query and each model, we perform five independent decoding runs under the same evaluation protocol. This results in a $5 \times 5$ observation matrix for each query-model pair, where each entry records both the response quality and the corresponding inference cost.
Our model pool includes Gemma-3-12B-IT~\citep{gemmateam2025gemma3technicalreport}, 
Mistral-Small-3.2-24B-Instruct~\citep{mistral2025small32}, 
Qwen3-32B~\citep{yang2025qwen3}, 
Llama-3.3-70B-Instruct~\citep{meta2024llama33}, 
Gemini-2.5-Flash-Lite~\citep{google2025gemini25flashlite}, 
and DeepSeek-Chat-V3.1~\citep{liu2024deepseek}.

\begin{observationbox}
\textbf{Observation 1: Single-shot labels are unstable across datasets.}
\end{observationbox}

Figure~\ref{fig:ana_instability} shows that single-shot supervision is unstable across all three datasets. We report three diagnostic metrics. \textbf{Outcome instability} measures whether repeated observations of the same query-model pair produce different scores. \textbf{Winner flip rate} measures whether the model selected as the best candidate for a query changes under different single-shot samples. \textbf{Single vs. DA disagreement} measures how often a single-shot routing label disagrees with the distribution-aware label estimated from repeated observations. 

The results show that instability is not limited to response scores, but propagates to model selection. GPQA exhibits the strongest instability, with an outcome instability of 0.715 and a winner flip rate of 0.970. DROP-800 also shows substantial instability, with a winner flip rate of 0.665 and a single-vs.-DA disagreement rate of 0.471. Even on MATH-500, where the task is more constrained, the winner flip rate remains 0.645. These results indicate that single-shot labels are not merely noisy measurements; they can directly change which model is selected for a query.

\begin{observationbox}
\textbf{Observation 2: Different tasks exhibit different uncertainty profiles.}
\end{observationbox}

Figure~\ref{fig:ana_uncertainty} decomposes uncertainty into input-side and output-side components. GPQA and MATH-500 are dominated by output-side uncertainty: their output-side variances are $0.099$ and $0.053$, respectively, clearly higher than their input-side variances. This indicates that repeated decoding alone can substantially change observed model behavior on scientific and mathematical reasoning tasks. In contrast, DROP-800 shows comparable input-side and output-side uncertainty, with input-side uncertainty slightly higher ($0.0428$ vs. $0.0350$). This suggests that reading-comprehension performance is also sensitive to how the query is formulated. Therefore, single-shot supervision can fail in different ways across tasks: some tasks are mainly affected by stochastic generation, while others are also sensitive to meaning-preserving input variations.

\begin{observationbox}
\textbf{Observation 3: Single-shot supervision induces unstable learned routers.}
\end{observationbox}

To test whether label instability affects router learning, we construct $100$ single-shot training sets by randomly sampling one observation from the $5\times5$ matrix for each query-model pair, and train one router on each sampled training set. Figure~\ref{fig:ana_router_variance} reports the resulting test performance distribution. Since all routers use the same model pool, data split, and architecture, the performance variation mainly comes from the random single-shot supervision used for training. The observed spread across the $100$ routers indicates that single-shot supervision can lead to non-negligible policy variance. Thus, the issue is not limited to label construction: stochastic supervision can propagate into the learned router and make the final model-selection policy less reliable.
\section{DARS: Distribution-Aware Routing Supervision}
\label{sec:DARS}

\begin{figure*}[t]
	\centering
	\includegraphics[width=0.8\linewidth]{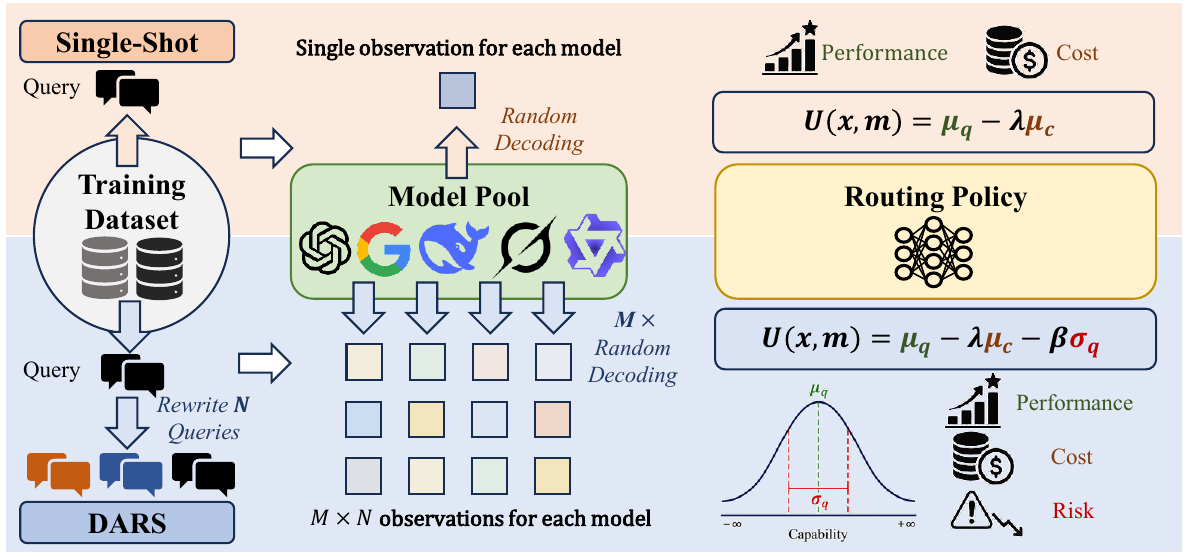}
    \caption{Overview of the proposed DARS framework. Conventional single-shot supervision obtains only one stochastic observation for each query-model pair and constructs routing supervision from the observed performance and cost. DARS instead collects multiple observations through query rewriting and repeated stochastic decoding, from which it estimates distributional capability signals, including expected performance, expected cost, and performance variability. These signals are then used to construct risk-aware supervision for learning a more reliable routing policy.}
    \label{fig:DARS}
\end{figure*}

\subsection{Overview of DARS}

LLM routing aims to select an appropriate model from a candidate pool \(\mathcal{M}\) for each input query \(x\), typically balancing response quality and inference cost. Existing routing methods often construct supervision from a single sampled response for each query-model pair, where the observed quality and cost are used to estimate a single-shot utility \(U_{\text{single}}(x,m)=q(x,m)-\lambda c(x,m)\). While simple, this point-estimate protocol is unreliable for stochastic LLM generation: semantically equivalent query formulations and repeated decoding runs may produce different responses, scores, and even model preferences. As a result, routers trained on single-shot observations may learn incidental sampling noise rather than stable differences in model capability.

To address this issue, we propose \textbf{DARS}, a distribution-aware supervision framework for LLM routing. As shown in Figure~\ref{fig:DARS}, DARS replaces isolated single-shot observations with repeated observations obtained through query rewriting and stochastic decoding. For each query-model pair, these observations are summarized into distributional capability signals, including expected performance, expected cost, and aggregate performance variability. DARS then constructs a risk-aware utility:
\[
U(x,m)
=
\mu_q(x,m)
-
\lambda \mu_c(x,m)
-
\beta \sigma_q(x,m),
\]
where \(\mu_q(x,m)\), \(\mu_c(x,m)\), and \(\sigma_q(x,m)\) denote the expected quality, expected cost, and overall performance variability, respectively. By incorporating both cost and risk into supervision, DARS encourages routers to select models that are not only accurate and cost-effective, but also stable under natural input and output variations.

\subsection{DARS Workflow}

DARS consists of four main stages: repeated observation construction, distributional capability estimation, risk-aware supervision construction, and routing policy learning.

\paragraph{Repeated observation construction.}

Given a training query \(x\), DARS first uses a strong language model to generate a set of semantically equivalent rewritten queries:
\[
\mathcal{X}(x)
=
\{x^{(1)}, x^{(2)}, \ldots, x^{(N)}\}.
\]
These rewrites preserve the original task semantics and gold answer, while varying the surface form, wording, or prompt structure. This step is intended to introduce controlled input-side perturbations. By observing how each candidate model behaves across these meaning-preserving variants, DARS can estimate whether the model's performance is stable with respect to different formulations of the same underlying query.

For each candidate model \(m \in \mathcal{M}\), DARS then applies stochastic decoding to the rewritten queries multiple times. This step captures output-side variability: even when the input formulation is fixed, different decoding runs may lead to different reasoning paths, final answers, or response lengths. For each generated response, DARS records its task-specific quality score and inference cost. Together with the rewritten-query dimension, these repeated decoding results form a set of observations for each query-model pair, which will be summarized into distributional capability signals in the next stage.

\paragraph{Distributional capability estimation.}

Let
\[
\mathcal{O}_{x,m}
=
\{(q_m^{(n,j)}, c_m^{(n,j)})\}
\]
denote the repeated observations collected for model \(m\) on query \(x\). DARS summarizes this observation set into three capability signals. The expected quality and expected cost are computed as
\[
\mu_q(x,m)
=
\operatorname{Mean}
\left(
\{q_m^{(n,j)}\}_{(n,j)}
\right),
\]
\[
\mu_c(x,m)
=
\operatorname{Mean}
\left(
\{c_m^{(n,j)}\}_{(n,j)}
\right).
\]
The performance risk is estimated by the standard deviation of the observed quality scores:
\[
\sigma_q(x,m)
=
\operatorname{Std}
\left(
\{q_m^{(n,j)}\}_{(n,j)}
\right).
\]
This risk term captures the aggregate variability caused by query rewriting and stochastic decoding. We use this overall variability as a practical risk signal for router training, rather than explicitly decomposing it into input-side and output-side components.

\paragraph{Risk-aware supervision construction.}

DARS combines the estimated capability signals into the following risk-aware utility:
\[
U(x,m)
=
\mu_q(x,m)
-
\lambda \mu_c(x,m)
-
\beta \sigma_q(x,m).
\]
The first term favors models with higher expected performance, the second term penalizes inference cost, and the third term penalizes unstable behavior. Based on this utility, the preferred model for query \(x\) is
\[
m^*(x)
=
\arg\max_{m \in \mathcal{M}}
U(x,m).
\]
Compared with single-shot supervision, this label is estimated from repeated observations and is therefore less sensitive to incidental sampled outcomes.

\paragraph{Routing policy learning.}

Finally, DARS provides distribution-aware supervision for learning the routing policy. Since DARS constructs model-wise capability signals rather than prescribing a specific router architecture, the resulting supervision can be adapted to a variety of learning paradigms.

For regression-based routers, DARS provides continuous targets such as expected quality, expected cost, and performance risk. The router can learn to predict these quantities for each candidate model and derive the final routing score from the predicted capability signals. For utility-based routers, the risk-aware utility \(U(x,m)\) can be directly used as the learning target, so that the router learns to approximate the final model-selection criterion. For classification-based routers, the model with the largest DARS utility can be used as a hard routing label, replacing the single-shot best-model label with a more stable distribution-aware label.

For probabilistic classification routers, DARS utilities can further be converted into a soft categorical target over the candidate model pool:
\[
p_{\text{DARS}}(m \mid x)
=
\frac{
\exp(U(x,m)/\tau)
}{
\sum_{m' \in \mathcal{M}}
\exp(U(x,m')/\tau)
},
\]
where \(\tau\) is a temperature parameter controlling the sharpness of the target distribution. This formulation treats model routing as a multi-class prediction problem, where candidate models are regarded as classes and the router is trained to match the DARS-induced model preference distribution. Compared with a hard label \(m^*(x)\), the soft target preserves relative preferences among models, especially when several candidates have similar risk-aware utilities.

DARS can also support ranking-based routers. In this case, the utility scores of candidate models are transformed into pairwise or listwise preferences, allowing the router to learn which model should be preferred for a given query. Similarly, for preference-based routing methods, DARS can provide more reliable preference labels by comparing models according to their distribution-aware utilities rather than single sampled outcomes. For non-parametric or retrieval-based routers, the estimated capability signals can be aggregated from similar training queries to infer the suitability of each candidate model for a new query. For cluster-based routers, DARS statistics can be summarized at the cluster level, enabling model selection based on the distributional behavior of queries with similar representations.

At inference time, the router selects the model with the highest predicted routing score, i.e., \(r(x)=\arg\max_{m \in \mathcal{M}} \hat{U}_{\theta}(x,m)\).

Overall, DARS decouples supervision construction from router design. By replacing noisy single-shot labels with distributional capability estimates, it provides a unified supervision source for regression-based, utility-based, classification-based, ranking-based, preference-based, retrieval-based, and cluster-based routing methods.

\section{Experiments}

\subsection{Experimental Setup}

\paragraph{Datasets and model pool.}
We evaluate DARS on the same datasets and model pool used in Section~\ref{sec:limitations_single_point}. The datasets cover different task formats and reasoning abilities, including scientific multiple-choice reasoning, mathematical problem solving, and reading comprehension. The candidate model pool contains heterogeneous LLMs with different capability and cost profiles. This setting allows us to examine whether distribution-aware supervision improves routing behavior across diverse tasks and model choices.

\paragraph{Test-time observation protocol.}
For evaluation, we construct two types of test observations for each query-model pair. First, we generate three semantically preserving rewrites for each test query to evaluate routing robustness under input-side variations. Second, we perform three stochastic decoding runs on the original query to evaluate robustness under output-side randomness. Based on these two observation groups, we report two test scores: rewrite-based utility, denoted as \(U_{\mathrm{rew}}\), and decoding-based utility, denoted as \(U_{\mathrm{dec}}\). The former measures whether the selected model remains effective when the query is expressed in different but equivalent forms, while the latter measures whether the selected model remains reliable under repeated stochastic generations.

\paragraph{Compared methods.}
We compare DARS with routing methods from different learning paradigms to evaluate whether distribution-aware supervision can benefit diverse router architectures. We consider MLPRouter \cite{stripelis2024tensoropera}, MIRT \cite{song2025irt}, and EmbedLLM \cite{zhuang2025embedllm} as regression-based routing methods; RM-Softmax \cite{tsiourvas2025causal} and GraphRouter \cite{feng2025graphrouter} as classification-based methods; AvengersPro \cite{zhang2025beyond} as a clustering-based method; and kNNRouter \cite{stripelis2024tensoropera} as a non-parametric retrieval-based method.

\paragraph{Router training protocol.}
For all baselines, DARS keeps the router architecture, query
representation, and training procedure unchanged, and only replaces
single-shot targets with distribution-aware supervision.
Regression-based routers learn model-wise capability statistics,
while classification-, retrieval-, and clustering-based routers use
the corresponding DARS-induced utilities or aggregated statistics.
Detailed implementations are provided in Appendix~\ref{app:baseline_implementation}.

\subsection{Experimental Results}
\paragraph{Overall comparison.}

Table~\ref{tab:DARS_reset_results} compares single-shot routing with DARS-enhanced routing across router families, with all utilities computed using \(\lambda=0.05\). For each baseline router, the single-shot result is averaged over 100 independently sampled training sets, while DARS keeps the router architecture unchanged and only replaces the supervision with distribution-aware signals constructed from repeated observations. This setup isolates the effect of supervision construction.

Overall, DARS improves routing utility in 40 of the 42
router--dataset--evaluation settings in Table~\ref{tab:DARS_reset_results}.
The gains are particularly evident on GPQA, where the diagnostic analysis
reveals the strongest single-shot instability, and are broadly observed
on DROP-800 and MATH-500.
The improvements span regression-, classification-, retrieval-, and
clustering-based routers, suggesting that the benefit of DARS is not tied
to a particular router design.

\begin{table}[t]
\centering
\caption{Comparison between single-shot routing and DARS-enhanced routing.}
\label{tab:DARS_reset_results}
\resizebox{\linewidth}{!}{
\begin{tabular}{lcccccc}
\toprule
\multirow{2}{*}{Router}
& \multicolumn{2}{c}{DROP-800}
& \multicolumn{2}{c}{GPQA}
& \multicolumn{2}{c}{MATH-500} \\
\cmidrule(lr){2-3}
\cmidrule(lr){4-5}
\cmidrule(lr){6-7}
& $U_{\mathrm{rew}}$ & $U_{\mathrm{dec}}$
& $U_{\mathrm{rew}}$ & $U_{\mathrm{dec}}$
& $U_{\mathrm{rew}}$ & $U_{\mathrm{dec}}$ \\
\midrule

MLP
& 0.661 & 0.709
& 0.382 & 0.400
& 0.682 & 0.698 \\
\rowcolor{DARSGray}
\quad + DARS
& \gain{0.675} & \gain{0.726}
& \gain{0.399} & \gain{0.413}
& \gain{0.687} & 0.696 \\

\cmidrule(lr){1-7}

kNN
& 0.713 & 0.759
& 0.441 & 0.469
& 0.720 & 0.733 \\
\rowcolor{DARSGray}
\quad + DARS
& \gain{0.749} & \gain{0.791}
& \gain{0.461} & \gain{0.489}
& \gain{0.730} & \gain{0.742} \\

\cmidrule(lr){1-7}

EmbedLLM
& 0.739 & 0.785
& 0.469 & 0.496
& 0.706 & 0.728 \\
\rowcolor{DARSGray}
\quad + DARS
& \gain{0.763} & \gain{0.805}
& \gain{0.473} & 0.480
& \gain{0.719} & \gain{0.747} \\

\cmidrule(lr){1-7}

AvengersPro
& 0.731 & 0.775
& 0.464 & 0.490
& 0.720 & 0.736 \\
\rowcolor{DARSGray}
\quad + DARS
& \gain{0.762} & \gain{\textbf{0.807}}
& \gain{0.480} & \gain{0.496}
& \gain{0.734} & \gain{0.742} \\

\cmidrule(lr){1-7}

GraphRouter
& 0.762 & 0.805
& 0.462 & 0.487
& 0.746 & 0.755 \\
\rowcolor{DARSGray}
\quad + DARS
& \gain{\textbf{0.765}} & \gain{\textbf{0.807}}
& \gain{\textbf{0.496}} & \gain{\textbf{0.516}}
& \gain{0.757} & \gain{0.758} \\

\cmidrule(lr){1-7}

MIRT
& 0.687 & 0.734
& 0.409 & 0.433
& 0.697 & 0.714 \\
\rowcolor{DARSGray}
\quad + DARS
& \gain{0.705} & \gain{0.751}
& \gain{0.444} & \gain{0.468}
& \gain{\textbf{0.759}} & \gain{\textbf{0.760}} \\

\cmidrule(lr){1-7}

RM-Softmax
& 0.740 & 0.787
& 0.453 & 0.475
& 0.721 & 0.746 \\
\rowcolor{DARSGray}
\quad + DARS
& \gain{0.761} & \gain{0.804}
& \gain{0.477} & \gain{0.512}
& \gain{0.731} & \gain{0.749} \\

\bottomrule
\end{tabular}
}
\end{table}

\begin{figure*}[t]
    \centering

    \begin{subfigure}[t]{0.32\linewidth}
        \centering
        \includegraphics[width=\linewidth]{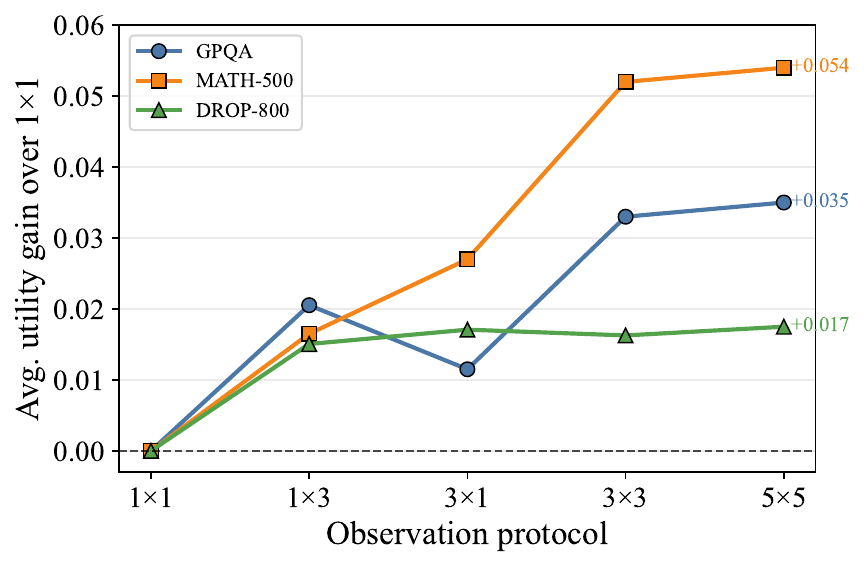}
        \caption{Sample efficiency of DARS.}
        \label{fig:sample_efficiency}
    \end{subfigure}
    \hfill
    \begin{subfigure}[t]{0.32\linewidth}
        \centering
        \includegraphics[width=\linewidth]{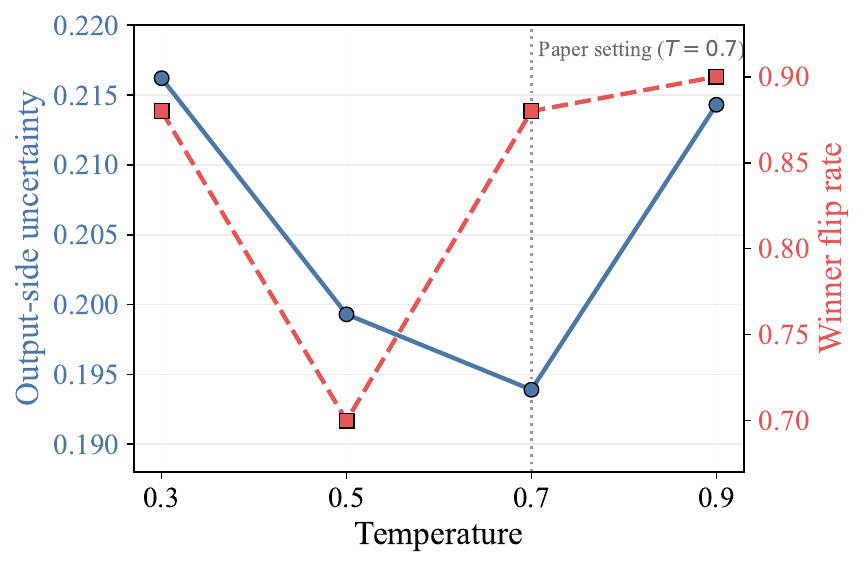}
        \caption{Sensitivity to the decoding temperature.}
        \label{fig:temperature_sensitivity}
    \end{subfigure}
    \hfill
    \begin{subfigure}[t]{0.32\linewidth}
        \centering
        \includegraphics[width=\linewidth]{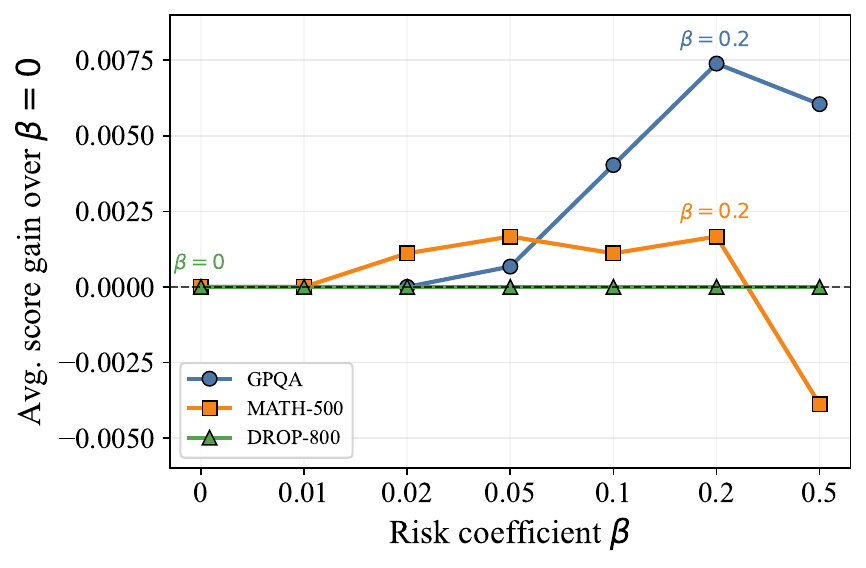}
        \caption{Sensitivity to the risk coefficient \(\beta\).}
        \label{fig:beta_sensitivity}
    \end{subfigure}

    \caption{
    Further analysis of DARS.
    (a) Routing utility under different observation protocols.
    (b) Output-side uncertainty and winner flip rate across decoding temperatures, with \(T=0.7\) used by default.
    (c) Sensitivity to the risk coefficient \(\beta\).
    }
    \label{fig:DARS_analysis}
\end{figure*}

\begin{figure}[t]
    \centering
    \includegraphics[width=0.92\linewidth]{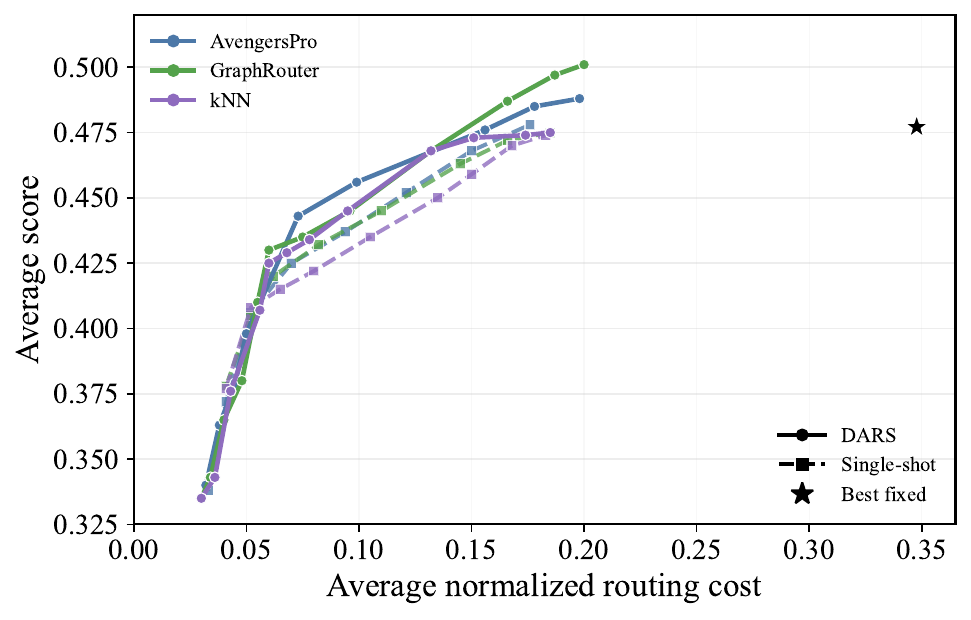}

    \caption{
    Cost--quality trade-offs under different routing budgets on GPQA.
    We compare DARS with single-shot supervision on three representative routers.
    }
    \label{fig:pareto_curve}
\end{figure}

\paragraph{Ablation study.}
Table~\ref{tab:DARS_ablation} isolates the effects of repeated observations and risk modeling.
Both Input-only DA and Output-only DA outperform single-shot supervision on average, showing that query reformulation and repeated decoding provide complementary supervision.
Removing the risk term also degrades performance, confirming the benefit of modeling performance variability beyond simply averaging repeated observations.

We further compare joint and decomposed risk modeling.
\textit{DARS w/ Decomp. Risk} estimates input- and output-side variability separately and combines the two penalties additively.
It improves the average utility from $0.626$ without risk modeling to $0.640$, but remains below Full DARS ($0.648$), which estimates variability over the joint observation set.
This suggests that the two uncertainty sources are not fully independent, since query reformulation can alter the generation distribution and thereby affect decoding variability.
We therefore use joint performance variability in DARS.

\begin{table}[t]
\centering
\small
\setlength{\tabcolsep}{4.0pt}
\renewcommand{\arraystretch}{1.08}
\caption{
Component ablation of DARS.
We compare distributional observations, risk modeling, and separate versus joint modeling of input- and output-side uncertainty.
}
\label{tab:DARS_ablation}
\resizebox{\linewidth}{!}{
\begin{tabular}{lccccccc}
\toprule
\multirow{2}{*}{Variant}
& \multicolumn{2}{c}{DROP-800}
& \multicolumn{2}{c}{GPQA}
& \multicolumn{2}{c}{MATH-500}
& \multirow{2}{*}{Avg.} \\
\cmidrule(lr){2-3}
\cmidrule(lr){4-5}
\cmidrule(lr){6-7}
& $U_{\mathrm{rew}}$ & $U_{\mathrm{dec}}$
& $U_{\mathrm{rew}}$ & $U_{\mathrm{dec}}$
& $U_{\mathrm{rew}}$ & $U_{\mathrm{dec}}$
&  \\
\midrule

Single-shot
& 0.687 & 0.734
& 0.409 & 0.433
& 0.697 & 0.714
& 0.612 \\

\midrule

Input-only DA
& 0.695 & 0.742
& 0.423 & 0.452
& 0.735 & 0.742
& 0.632 \\

Output-only DA
& 0.693 & 0.747
& 0.425 & 0.442
& 0.723 & 0.718
& 0.625 \\

\midrule

DARS w/o Risk
& 0.692 & 0.731
& 0.413 & 0.441
& 0.746 & 0.730
& 0.626 \\

DARS w/ Decomp. Risk
& 0.705 & 0.743
& 0.435 & 0.456
& 0.754 & 0.749
& 0.640 \\

\rowcolor{DARSGray}
Full DARS
& \textbf{0.705} & \textbf{0.751}
& \textbf{0.444} & \textbf{0.468}
& \textbf{0.759} & \textbf{0.760}
& \textbf{0.648} \\

\bottomrule
\end{tabular}
}
\end{table}

\paragraph{Sample efficiency of DARS.}
Figure~\ref{fig:sample_efficiency} studies the effect of different observation protocols.
Increasing the number of observations generally improves routing utility, but the gains quickly saturate.
In particular, the \(3\times3\) protocol already performs close to the full \(5\times5\) setting across datasets, showing that DARS does not require exhaustive sampling to obtain reliable supervision.

We further evaluate an adaptive sampling strategy that starts from a \(2\times2\) observation grid and expands the budget only when the estimated risk remains high or the utility margin between the top two candidate models is small.
On GPQA, 50 of the 200 training queries stop at \(2\times2\), while the remaining 150 expand to \(3\times3\), resulting in only \(7.75\) observations per query--model pair on average.
This corresponds to a \(13.9\%\) reduction in candidate-model calls relative to fixed \(3\times3\) sampling.
These results show that the supervision cost of DARS can be further reduced by allocating additional observations primarily to ambiguous queries.

\paragraph{Sensitivity to decoding temperature.}
Figure~\ref{fig:temperature_sensitivity} examines whether the observed output-side instability depends on the decoding temperature.
Both output-side uncertainty and winner flip rates remain substantial across all tested temperatures.
Although their exact values vary with temperature, neither exhibits a monotonic trend, and the default setting \(T=0.7\) does not produce the largest uncertainty.
This indicates that the instability of single-shot supervision is not an artifact of a particular decoding temperature.

\paragraph{Sensitivity to the risk coefficient.}
Figure~\ref{fig:beta_sensitivity} shows that DARS is generally stable across a broad range of risk coefficients.
Among the tested values, \(\beta=0.2\) provides a strong and stable setting across datasets, and we use it as the default in the main experiments.

\paragraph{Cost--quality trade-off under different budgets.}
Figure~\ref{fig:pareto_curve} compares DARS with single-shot supervision under different routing budgets on GPQA.
Across the representative routers, DARS generally shifts the cost--quality frontier upward, achieving higher quality at comparable cost or similar quality at lower cost.
DARS-enhanced routers also surpass the best fixed-model baseline in several cost regions.
Thus, the benefit of distribution-aware supervision extends beyond average routing utility to practical cost--quality trade-offs.
\section{Conclusion}

In this paper, we identify the limitation of single-shot supervision in LLM routing and propose DARS, a distribution-aware supervision framework that estimates model capability from repeated observations across query rewrites and stochastic decoding. Experiments show that DARS provides more reliable routing signals and generally improves routing performance across diverse router families under cost-quality-risk trade-offs.

\section*{Limitations}

This work primarily studies single-step LLM routing with a fixed
candidate model pool. Extending distribution-aware supervision to
multi-round or agentic routing is an interesting direction, where
capability estimates may need to be defined over intermediate states
or interaction trajectories rather than individual queries.

DARS also requires repeated observations to construct supervision,
which introduces additional offline data-collection cost compared with
single-shot training. Our sample-efficiency and adaptive-sampling
results suggest that this overhead can be reduced with more efficient
observation allocation. Future work may further explore adaptive
sampling strategies and extend DARS to broader routing settings and
model pools.

\section*{Ethical considerations}

This work studies model routing and supervision construction for selecting among existing LLMs. It does not introduce new user-facing generation objectives or collect private user data. The main goal is to improve the reliability and cost-effectiveness of LLM routing by constructing more stable supervision signals.

However, routing systems may affect which model is used for a given query, and different models may have different safety, bias, privacy, and reliability characteristics. In practical deployment, DARS should therefore be used together with appropriate model-level safety filters, privacy protections, and monitoring mechanisms. Since routing also involves inference cost, systems based on DARS should maintain transparent cost accounting and avoid selecting unnecessarily expensive models when cheaper models are sufficient.

\section*{Acknowledgments}
This work was supported by the National Key R\&D Program of China under Grant No. 2024YFE0202800 and the National Natural Science Foundation of China under Grant Nos. 62522605 and 62376118. Long Chen was additionally supported by the National Natural Science Foundation of China under Grant Nos. 62522216 and 62402408, and the Hong Kong SAR Research Grants Council under Grant Nos. 26208924 and 16219025.

\bibliography{custom}

\appendix

\section*{Appendix}
\label{sec:appendix}

\section{Dataset and Model Details}
\label{app:dataset_model_details}

\paragraph{Datasets.}
We evaluate DARS on three datasets that cover different task formats and reasoning requirements. 
\textbf{GPQA} is a graduate-level multiple-choice question answering benchmark covering challenging scientific domains such as biology, physics, and chemistry \citep{rein2024gpqa}. 
\textbf{MATH-500} is a 500-problem subset of the MATH benchmark, focusing on mathematical problem solving and multi-step reasoning \citep{hendrycks2021measuring, lightman2024let}. 
\textbf{DROP-800} is a subset of the DROP reading-comprehension benchmark, where questions require discrete reasoning over paragraphs, such as counting, sorting, comparison, and arithmetic reasoning \citep{dua2019drop}. 
These datasets are selected because they differ in output format, reasoning style, and sensitivity to input/output variations, allowing us to evaluate whether distribution-aware supervision is useful across heterogeneous routing scenarios.

\paragraph{Data split.}
For each dataset, we construct a training split for learning routing policies and a held-out test split for evaluation. 
Following the diagnostic analysis in the main paper, we sample 200 training queries from each dataset to construct repeated observations for supervision. 
The remaining held-out queries are used for test-time evaluation. 
All routers are trained and evaluated using the same split to ensure that performance differences come from the supervision construction method rather than from different data partitions.

\paragraph{Training observation protocol.}
For each training query, we construct distribution-aware observations for every candidate model. 
To capture input-side variation, we use a strong language model to generate five semantically preserving rewrites of the original query. 
The rewriting process preserves the original task semantics, gold answer, and candidate choices when applicable. 
To capture output-side stochasticity, each candidate model is evaluated with five independent decoding runs for each rewritten query. 
This gives a \(5 \times 5\) observation matrix for every query-model pair, where each entry records the response quality and the corresponding inference cost. 
DARS summarizes these repeated observations into expected quality, expected cost, and performance variability, which are then used to construct distribution-aware supervision signals.

\paragraph{Single-shot training protocol.}
For the single-shot baseline, each query-model pair is represented by one sampled observation. 
To reduce the effect of a particular random draw, we construct 100 independently sampled single-shot training sets. 
Each single-shot training set contains one randomly selected observation for every query-model pair, and the reported single-shot result is averaged over the 100 trained routers. 
In contrast, the DARS-enhanced version keeps the router architecture unchanged and replaces only the single-shot supervision with distribution-aware supervision constructed from repeated observations.

\paragraph{Test-time evaluation protocol.}
At test time, the router receives only the original query and selects one model from the candidate pool. 
Repeated test observations are used only to evaluate the selected model, not to provide additional information to the router. 
For each test query, we construct two evaluation groups. 
The first group contains three semantically preserving rewrites of the query, which is used to compute the rewrite-based utility \(U_{\mathrm{rew}}\). 
The second group contains three independent stochastic decoding runs on the original query, which is used to compute the decoding-based utility \(U_{\mathrm{dec}}\). 
The former measures whether the routing decision is robust to input-side variations, while the latter measures whether the selected model remains reliable under output-side randomness.

\paragraph{Model pool.}
We use a heterogeneous pool of six LLMs with different capability and cost profiles: Gemma-3-12B-IT \citep{gemmateam2025gemma3technicalreport}, Mistral-Small-3.2-24B-Instruct \citep{mistral2025small32}, Qwen3-32B \citep{yang2025qwen3}, Llama-3.3-70B-Instruct \citep{meta2024llama33}, Gemini-2.5-Flash-Lite \citep{google2025gemini25flashlite}, and DeepSeek-Chat-V3.1 \citep{liu2024deepseek}. 
This model pool contains both open-source and API-based models, as well as models with different sizes and deployment costs. 
Such heterogeneity is important for evaluating LLM routing, since the router must learn when a cheaper model is sufficient and when a stronger model is worth the additional cost.

\paragraph{Evaluation metric and cost setting.}
For each generated response, we compute a task-specific quality score and an inference cost. 
Quality is measured by the corresponding task metric, such as accuracy for multiple-choice and mathematical tasks, and F1-style matching for reading-comprehension tasks. 
Inference cost is normalized within each dataset before computing routing utilities. 
Unless otherwise specified, we use the cost coefficient \(\lambda=0.05\) and the risk coefficient \(\beta=0.2\) in the main experiments.

\section{Query Rewriting Prompt and Decoding Protocol}
\label{app:rewriting_decoding}

\paragraph{Query rewriting.}
For each training query, we generate semantically preserving rewrites using a strong language model. The purpose of rewriting is to introduce controlled input-side variation while keeping the underlying task unchanged. We use dataset-specific rewriting instructions to avoid changing the gold answer, answer choices, mathematical conditions, or passage-grounded evidence. The rewriting constraints are shown below.

For GPQA, only the question stem is rewritten, while the answer choices are kept unchanged. For MATH-500, all mathematical constraints and requested quantities must be preserved. For DROP-800, only the question is rewritten and the passage remains fixed. These constraints ensure that each rewrite remains a valid variant of the original query and can be evaluated using the same gold answer.

\begin{tcolorbox}[
    colback=gray!5,
    colframe=gray!60,
    title=Dataset-specific rewriting instructions,
    fonttitle=\bfseries,
    boxrule=0.5pt,
    arc=2pt
]
\small
\textbf{GPQA.} This is a graduate-level multiple-choice science question. Rewrite only the question stem. Do not modify, reorder, remove, or paraphrase the answer choices. The correct option after rewriting must remain exactly the same.

\vspace{0.5em}

\textbf{MATH-500.} This is a free-form mathematics problem. Rewrite only the problem statement. Preserve every mathematical condition, number, variable, equation, unit, constraint, and requested quantity. The final mathematical answer must remain exactly the same. Do not simplify the problem, add hints, or reveal the answer.

\vspace{0.5em}

\textbf{DROP-800.} This is a reading-comprehension question over a fixed passage. Rewrite only the question. Do not rewrite the passage. The rewritten question must be answerable from the same passage and must have the same gold answer. Do not add information that is not supported by the passage.
\end{tcolorbox}

\paragraph{Prompt format for model generation.}
After obtaining rewritten queries, we construct dataset-specific prompts for candidate models. All prompts require the model to output a clearly marked \texttt{Final Answer:} line, so that responses can be parsed and evaluated consistently. For GPQA, the prompt includes the rewritten question and the original answer choices, and asks the model to return one option letter. For MATH-500, the prompt asks the model to solve the rewritten mathematics problem and put only the final result after \texttt{Final Answer:}. For DROP-800, the prompt provides the fixed passage and the rewritten question, and asks the model to answer using only the passage.

\begin{tcolorbox}[
    colback=gray!5,
    colframe=gray!60,
    title=Shared system instruction for generation,
    fonttitle=\bfseries,
    boxrule=0.5pt,
    arc=2pt
]
\small
You are answering benchmark questions for an LLM routing experiment. Follow the user instructions exactly. You may reason internally, but the final response must contain a clearly marked \texttt{Final Answer:} line. Do not mention that this is a paraphrase or dataset example.
\end{tcolorbox}

\paragraph{Training decoding protocol.}
For each training query, DARS uses rewritten prompt variants rather than the original query by default. Specifically, we use \(N=5\) rewritten variants for each query. For every rewritten variant and every candidate model, we perform \(M=5\) independent stochastic decoding runs. This produces \(5 \times 5\) observations for each query-model pair. Each observation records the generated response, task-specific quality score, token usage, and inference cost. These observations are then aggregated into expected quality, expected cost, and performance variability for constructing DARS supervision.

\paragraph{Decoding configuration.}
All model generations are collected through the same decoding interface using the same sampling configuration unless otherwise specified. We use temperature \(0.7\) and top-p \(0.95\) to induce stochastic generations. The maximum generation length is capped according to the dataset: 768 tokens for GPQA, 3072 tokens for MATH-500, and 512 tokens for DROP-800. Independent decoding samples are obtained by issuing separate generation calls with the same prompt and decoding configuration. We use the same candidate model pool as described in Appendix~\ref{app:dataset_model_details}.

\paragraph{Test-time decoding protocol.}
At test time, the router receives only the original query and selects a candidate model. The repeated test observations are used only for evaluation and are not provided to the router. For each selected model, we compute two evaluation views. The rewrite-based utility \(U_{\mathrm{rew}}\) is evaluated using three semantically preserving rewrites of the test query, while the decoding-based utility \(U_{\mathrm{dec}}\) is evaluated using three stochastic decoding runs on the original query. This protocol allows us to separately evaluate robustness to input-side query variation and output-side generation randomness.

\section{Supplementary Experimental Details and Results}

\subsection{Baseline Implementation Details}
\label{app:baseline_implementation}

This section provides implementation details for the routing baselines used in our experiments. All baselines are implemented in the same codebase and share the same data loader, query feature extraction pipeline, candidate model pool, supervision construction protocol, and evaluation procedure unless otherwise specified.

\paragraph{Common setup.}
All methods are evaluated on GPQA, MATH-500, and DROP-800 using the same six-model candidate pool described in Appendix~\ref{app:dataset_model_details}. Each method reads scored generations from the same training and test files. For query representation, we use one feature vector per query. By default, the feature text contains both the question and available context fields; context can be removed for ablation. Neural and retrieval-based methods use the same query encoder interface. When available, we use a local sentence-transformer encoder; otherwise, we fall back to TF-IDF features. The default TF-IDF dimensionality is 20,000.

\paragraph{Single-shot and DARS supervision.}
Each baseline is evaluated under two supervision modes. In the single-shot setting, each query-model pair is represented by one sampled scored observation. To reduce the effect of a particular random draw, we repeat this process with 100 independently sampled single-shot training sets and report the averaged result. In the DARS setting, the router architecture is kept unchanged, but the supervision signal is replaced by distribution-aware statistics aggregated from repeated observations. Specifically, for each query-model pair, we compute the mean score, mean cost, and score standard deviation. The default inference utility is based on the risk-aware objective \( \text{score} - \lambda \cdot \text{cost} - \beta \cdot \text{risk} \), where the cost coefficient is set to \(\lambda=0.05\).
\begin{table*}[t]
\centering
\small
\setlength{\tabcolsep}{4pt}
\renewcommand{\arraystretch}{1.08}
\caption{Summary of baseline implementations and how DARS supervision is incorporated.}
\label{tab:baseline_implementation}
\resizebox{\linewidth}{!}{
\begin{tabular}{lll}
\toprule
Router & Learning paradigm & DARS-enhanced supervision \\
\midrule
MLP & Regression-based routing & Predicts mean score, mean cost, and score standard deviation. \\
MIRT & IRT-style regression & Learns score, cost, and uncertainty through separate IRT-style heads. \\
EmbedLLM & Model-embedding regression & Uses risk-adjusted performance targets with learned model embeddings. \\
RM-Softmax & Classification / regret minimization & Trains on risk-adjusted utilities across cost coefficients. \\
GraphRouter & Query-model edge prediction & Predicts distribution-aware edge-level score, cost, and uncertainty. \\
kNNRouter & Non-parametric retrieval & Aggregates distribution-aware targets from nearest training queries. \\
AvengersPro & Cluster-based routing & Aggregates DARS statistics at the query-cluster level. \\
\bottomrule
\end{tabular}
}
\end{table*}
\paragraph{Test views.}
Each trained router predicts one model for each test query. The selected model is evaluated under two views. The rewrite view uses observations generated from semantically preserving query rewrites, while the decoding view uses repeated stochastic decoding samples from the original query. These test observations are used only for evaluation and are not provided to the router during model selection.

\paragraph{Regression-based routers.}
\textbf{MLP} trains separate regressors for model score, model cost, and, under DARS, score standard deviation. Each regressor predicts a vector over the candidate models for a given query representation, and the router selects the model with the largest predicted risk-adjusted utility. We use scikit-learn MLP regressors with standardized input features.

\textbf{MIRT} implements a multi-dimensional item-response-theory router. It maps each query into a latent ability vector and learns model-specific discrimination and difficulty parameters. Separate prediction heads are used for score, cost, and score standard deviation. Under DARS, MIRT learns all three distributional targets and selects models according to the predicted risk-aware utility.

\textbf{EmbedLLM} learns compact model embeddings and a query projection to predict query-model performance. It jointly trains a performance head and a model-cost head. Under DARS, the score target is adjusted using the performance-risk term, while the implementation does not explicitly predict query-specific uncertainty at inference time; the risk information is incorporated through the training target.

\paragraph{Classification-based routers.}
\textbf{RM-Softmax} implements a regret-minimization router. It augments each query embedding with a cost-weight value and trains a neural classifier over candidate models across a grid of cost coefficients. Under DARS, the target performance is risk-adjusted before training, so the classifier learns model preferences induced by distribution-aware supervision.

\textbf{GraphRouter} treats each query-model pair as an edge between a query node and a model node. Query features and model descriptions are projected into a shared space, and an edge predictor estimates model-wise score, cost, and uncertainty. Under DARS, these targets correspond to the aggregated mean score, mean cost, and score standard deviation for each query-model pair.

\paragraph{Non-parametric and cluster-based routers.}
\textbf{kNNRouter} is a non-parametric retrieval-based router. It stores training query embeddings and their query-model target matrices. For a test query, it retrieves the nearest training queries and averages their model-wise score, cost, and uncertainty statistics using inverse-distance weights. Under DARS, the stored targets are distribution-aware statistics rather than single-shot observations.

\textbf{AvengersPro} is implemented as a cluster-based router. Training queries are clustered in feature space, and each cluster stores the average score, cost, and uncertainty for each candidate model. For a new query, the router retrieves the nearest clusters and estimates model utility by aggregating their stored statistics. Under DARS, these cluster-level statistics are computed from distribution-aware query-model observations.

\subsection{Additional Experimental Results}
\label{app:additional_results}

Table~\ref{tab:additional_main_results} provides an expanded version of the main comparison table. 
For the single-shot setting, we report the mean and standard deviation over 100 independently sampled single-shot training sets when run-level standard deviations are available. 
For DARS, each router is trained once using distribution-aware supervision constructed from repeated observations. 
The results show that single-shot supervision can exhibit non-negligible variance across random sampled training sets, especially on GPQA. 
In contrast, replacing single-shot labels with DARS supervision consistently improves the mean performance of most routers, confirming that the gains in the main table come from more reliable supervision construction rather than changes to the router architecture.

\begin{table*}[t]
\centering
\small
\setlength{\tabcolsep}{4pt}
\renewcommand{\arraystretch}{1.08}
\caption{
Extended comparison between single-shot routing and DARS-enhanced routing.
For single-shot routing, values are averaged over 100 independently sampled training sets, and we report mean \(\pm\) standard deviation when run-level standard deviations are logged.
}
\label{tab:additional_main_results}
\begin{tabular}{llcccc}
\toprule
Router & Dataset
& Single \(U_{\mathrm{rew}}\)
& DARS \(U_{\mathrm{rew}}\)
& Single \(U_{\mathrm{dec}}\)
& DARS \(U_{\mathrm{dec}}\) \\
\midrule

\multirow{3}{*}{MLP}
& DROP-800 & \(0.661 \pm 0.009\) & 0.675 & \(0.709 \pm 0.011\) & 0.726 \\
& GPQA     & \(0.382 \pm 0.020\) & 0.399 & \(0.400 \pm 0.023\) & 0.413 \\
& MATH-500 & \(0.682 \pm 0.013\) & 0.687 & \(0.698 \pm 0.013\) & 0.696 \\
\midrule

\multirow{3}{*}{KNNRouter}
& DROP-800 & 0.713 & 0.749 & 0.759 & 0.791 \\
& GPQA     & 0.441 & 0.461 & 0.469 & 0.489 \\
& MATH-500 & 0.720 & 0.730 & 0.733 & 0.742 \\
\midrule

\multirow{3}{*}{EmbedLLM}
& DROP-800 & \(0.739 \pm 0.023\) & 0.763 & \(0.785 \pm 0.020\) & 0.805 \\
& GPQA     & \(0.469 \pm 0.022\) & 0.473 & \(0.496 \pm 0.021\) & 0.480 \\
& MATH-500 & \(0.706 \pm 0.019\) & 0.719 & \(0.728 \pm 0.022\) & 0.747 \\
\midrule

\multirow{3}{*}{AvengersPro}
& DROP-800 & 0.731 & 0.762 & 0.775 & 0.807 \\
& GPQA     & 0.464 & 0.480 & 0.490 & 0.496 \\
& MATH-500 & 0.720 & 0.734 & 0.736 & 0.742 \\
\midrule

\multirow{3}{*}{GraphRouter}
& DROP-800 & \(0.762 \pm 0.010\) & 0.765 & \(0.805 \pm 0.007\) & 0.807 \\
& GPQA     & \(0.462 \pm 0.025\) & 0.496 & \(0.487 \pm 0.022\) & 0.516 \\
& MATH-500 & \(0.746 \pm 0.013\) & 0.757 & \(0.755 \pm 0.007\) & 0.758 \\
\midrule

\multirow{3}{*}{MIRT}
& DROP-800 & 0.687 & 0.705 & 0.734 & 0.751 \\
& GPQA     & 0.409 & 0.444 & 0.433 & 0.468 \\
& MATH-500 & 0.697 & 0.759 & 0.714 & 0.760 \\
\midrule

\multirow{3}{*}{RM-Softmax}
& DROP-800 & \(0.740 \pm 0.012\) & 0.761 & \(0.787 \pm 0.013\) & 0.804 \\
& GPQA     & \(0.453 \pm 0.016\) & 0.477 & \(0.475 \pm 0.018\) & 0.512 \\
& MATH-500 & \(0.721 \pm 0.004\) & 0.731 & \(0.746 \pm 0.004\) & 0.749 \\

\bottomrule
\end{tabular}

\end{table*}

\subsection{Ablation Details}
\label{app:ablation_details}

This section describes how each ablation variant in Table~\ref{tab:DARS_ablation} is constructed. All variants use the same router architecture, training split, candidate model pool, and evaluation protocol. The only difference is how the supervision signal is constructed from repeated observations.

\paragraph{Single-shot.}
The single-shot baseline follows the standard routing supervision protocol. For each query-model pair, we randomly sample one scored observation from the available observation set and use it as the training target. Since the target contains only one observed score and cost, no performance-variability term is used. To reduce dependence on a particular random draw, we repeat this process with 100 independently sampled single-shot training sets and report the averaged result.

\paragraph{Input-only DA.}
This variant isolates the effect of query rewriting. For each query-model pair, we aggregate observations across rewritten queries, but do not use repeated decoding samples for each rewrite. In practice, this variant estimates model behavior from meaning-preserving input variations, so the resulting supervision captures sensitivity to query formulation while excluding output-side stochasticity as much as possible.

\paragraph{Output-only DA.}
This variant isolates the effect of stochastic decoding. Instead of aggregating over rewritten queries, it uses repeated decoding samples from the original query. The supervision signal therefore captures how model performance changes under generation randomness while keeping the input formulation fixed.

\paragraph{DARS w/o Risk.}
This variant uses the same repeated observations as full DARS, but removes the performance-variability term from the utility. It constructs supervision only from expected quality and expected cost. This ablation tests whether the benefit of DARS comes only from averaging repeated observations, or whether explicitly modeling instability provides additional value.

\paragraph{DARS w/ Decomposed Risk.}
This variant explicitly estimates input-side and output-side
performance variability separately and combines the two risk terms
additively. Let $\sigma_{\mathrm{in}}(x,m)$ and
$\sigma_{\mathrm{out}}(x,m)$ denote the variability induced by
query rewriting and stochastic decoding, respectively. The
decomposed-risk utility is

\begin{align}
    U_{\mathrm{decomp}}(x,m)
=
\mu_q(x,m)-\lambda\mu_c(x,m)
- \nonumber\\ 
\beta\left(
\sigma_{\mathrm{in}}(x,m)+
\sigma_{\mathrm{out}}(x,m)
\right).\nonumber
\end{align}

Unlike Full DARS, which directly estimates variability from the
joint observation set, this variant assumes that the two sources
can be modeled separately and combined additively.

\paragraph{Full DARS.}
Full DARS aggregates observations across both rewritten queries and repeated stochastic decoding runs. It estimates expected quality, expected cost, and aggregate performance variability from the complete observation set. This variant therefore captures both input-side and output-side uncertainty, and uses the resulting risk-aware utility to construct the final supervision signal.

\subsection{Utility Computation and Evaluation Details}
\label{app:utility_details}

This section provides additional details on how routing utilities and evaluation scores are computed.

\paragraph{Per-observation utility.}
Each generated response is associated with a task-specific quality score and an inference cost. Let \(q_{i,m}^{(o)}\) and \(c_{i,m}^{(o)}\) denote the quality score and normalized cost of model \(m\) on query \(x_i\) under observation \(o\). The per-observation utility is computed as
\[
u_{i,m}^{(o)}
=
q_{i,m}^{(o)}
-
\lambda c_{i,m}^{(o)}.
\]
In all main experiments, we set \(\lambda=0.05\).

\paragraph{Rewrite-based utility.}
The rewrite-based utility evaluates whether the selected model remains effective under semantically equivalent query formulations. For each test query, the router receives only the original query and selects one model \(r(x_i)\). We then evaluate the selected model on the rewrite-view observations, which contain three rewritten versions of the test query. The rewrite-based utility \(U_{\mathrm{rew}}\) is the average per-observation utility of the selected model over these rewrite-view observations and all test queries:

{\footnotesize
\[
U_{\mathrm{rew}}(r)
=
\frac{\sum_{x_i \in \mathcal{D}_{\mathrm{test}}}
\frac{\sum_{o \in \mathcal{O}^{\mathrm{rew}}_i}
\left(
q_{i,r(x_i)}^{(o)}
-
\lambda c_{i,r(x_i)}^{(o)}
\right)}{|\mathcal{O}^{\mathrm{rew}}_i|}
}{|\mathcal{D}_{\mathrm{test}}|}
,
\]
}
where \(\mathcal{O}^{\mathrm{rew}}_i\) denotes the rewrite-view observations of query \(x_i\).

\paragraph{Decoding-based utility.}
The decoding-based utility evaluates whether the selected model remains reliable under stochastic generation. The router again receives only the original query and selects one model \(r(x_i)\). We then evaluate this selected model on the decoding-view observations, which contain three independent stochastic decoding runs on the original query. The decoding-based utility \(U_{\mathrm{dec}}\) is computed as

{\footnotesize
\[
U_{\mathrm{dec}}(r)
=
\frac{\sum_{x_i \in \mathcal{D}_{\mathrm{test}}}
\frac{\sum_{o \in \mathcal{O}^{\mathrm{dec}}_i}
\left(
q_{i,r(x_i)}^{(o)}
-
\lambda c_{i,r(x_i)}^{(o)}
\right)}{|\mathcal{O}^{\mathrm{dec}}_i|}
}{|\mathcal{D}_{\mathrm{test}}|}
,
\]
}
where \(\mathcal{O}^{\mathrm{dec}}_i\) denotes the decoding-view observations of query \(x_i\).

\paragraph{Cost normalization.}
Raw inference costs can have different scales across datasets, models, and generation lengths. To make the cost term comparable with the quality score, we normalize costs before computing utilities. Let \(c^{\mathrm{raw}}_{i,m,o}\) denote the raw inference cost recorded for an observation in dataset \(d\). We use a dataset-level normalization:
\[
c_{i,m}^{(o)}
=
\frac{
c^{\mathrm{raw}}_{i,m,o}
}{
Z_d
},
\quad
Z_d
=
\max_{i,m,o}
c^{\mathrm{raw}}_{i,m,o}.
\]
After normalization, the cost values are on a comparable scale across models within the same dataset. All reported utilities use normalized costs.

\paragraph{Choice of \(\lambda\).}
The cost coefficient \(\lambda\) controls the trade-off between response quality and inference cost. We use \(\lambda=0.05\) for all main experiments. This value gives a moderate cost penalty after normalization: it is large enough to discourage unnecessarily expensive model choices, while avoiding a degenerate setting where the router always prefers the cheapest model. We keep \(\lambda\) fixed across datasets and routers so that differences in performance reflect the routing method and supervision signal rather than per-method tuning.

\paragraph{Why not model cost variability?}
DARS explicitly models performance variability while using the expected
inference cost in its default utility. Under our cost definition,
variation within a fixed query--model pair mainly arises from differences
in completion length, whereas the dominant cost differences for routing
come from the price and token usage differences across candidate models.
In contrast, performance can change qualitatively across repeated
generations, including transitions between correct and incorrect
outcomes, which can directly alter model preferences.
We therefore focus the risk term on performance variability and use
expected cost to capture the primary cost differences among candidate
models. Explicitly modeling cost variability may become more important
in settings with long-form generation, latency constraints, or strict
tail-cost budgets, which we leave for future work.

\paragraph{Best fixed baseline.}
The best fixed baseline represents the strongest non-routing strategy. It selects one single model and uses it for all test queries. For each dataset and evaluation view, we compute the average utility of every candidate model over all test queries, and report the best-performing fixed model:

{\footnotesize
\[
m_{\mathrm{fixed}}^*
=
\arg\max_{m \in \mathcal{M}}
\frac{
\sum_{x_i \in \mathcal{D}_{\mathrm{test}}}
\frac{\sum_{o \in \mathcal{O}_i}
\left(
q_{i,m}^{(o)}
-
\lambda c_{i,m}^{(o)}
\right)}{|\mathcal{O}_i|}
}{|\mathcal{D}_{\mathrm{test}}|}.
\]
}

This baseline is useful for evaluating whether a learned router provides a better cost-quality trade-off than always using the same strong model.

\paragraph{Use of test observations.}
Repeated test observations are used only for evaluation. At inference time, the router receives the original query and predicts one model. The rewrite-view and decoding-view observations are not provided to the router and are not used to revise its decision. This ensures that \(U_{\mathrm{rew}}\) and \(U_{\mathrm{dec}}\) evaluate the robustness of the selected model rather than giving the router additional test-time information.

\subsection{Efficiency and Sensitivity Analyses}
\label{app:efficiency_sensitivity}

This section provides additional details for the efficiency and
sensitivity analyses in Section~5.2, including adaptive observation
allocation, decoding-temperature sensitivity, and sensitivity to the
risk coefficient.

\paragraph{Adaptive sampling.}
A fixed observation budget allocates the same amount of computation to
every training query, although some queries already exhibit a stable and
well-separated model preference after only a few observations.
We therefore consider an adaptive acquisition strategy that progressively
expands the observation grid according to the estimated uncertainty and
the utility margin between the two most competitive candidate models.

For a query \(x\), we define the maximum estimated risk as
\begin{equation}
R(x)
=
\max_{m\in\mathcal{M}}
\left(
\hat{\sigma}_{\mathrm{in}}(x,m)
+
\hat{\sigma}_{\mathrm{out}}(x,m)
\right),
\end{equation}
where \(\hat{\sigma}_{\mathrm{in}}(x,m)\) and
\(\hat{\sigma}_{\mathrm{out}}(x,m)\) denote the current estimates of
input-side and output-side performance variability, respectively.
We also define the utility margin between the two highest-ranked
candidate models as
\begin{equation}
\Delta(x)
=
\hat{U}_{(1)}(x)-\hat{U}_{(2)}(x),
\end{equation}
where \(\hat{U}_{(1)}(x)\) and \(\hat{U}_{(2)}(x)\) are the largest
and second-largest estimated utilities under the current observation
set.
A small \(R(x)\) indicates relatively stable model behavior, whereas a
large \(\Delta(x)\) indicates that the leading candidate is clearly
separated from the second-best model.

For the GPQA adaptive experiment, sampling starts from a
\(2\times2\) observation grid, consisting of two query rewrites and two
independent stochastic decoding runs per rewrite, for a total of four
observations per query--model pair.
At this stage, sampling stops when
\begin{equation}
R(x)\leq Q_{0.25}(R)
\qquad\text{and}\qquad
\Delta(x)\geq Q_{0.75}(\Delta),
\end{equation}
where \(Q_p(\cdot)\) denotes the \(p\)-th quantile of the corresponding
query-level statistic computed over the training set.
Queries that do not satisfy both conditions are expanded to a
\(3\times3\) observation grid.

The acquisition procedure also permits a final expansion to a
\(5\times5\) grid for queries that remain ambiguous after the
\(3\times3\) stage.
At the second decision point, the stopping thresholds are set to
\(Q_{1.00}(R)\) for risk and \(Q_{0.00}(\Delta)\) for the utility
margin.
Under this GPQA configuration, all remaining queries therefore stop at
the \(3\times3\) stage, and the \(5\times5\) maximum budget is not
activated.
When the observation grid is expanded, all previously collected
observations are reused and the risk and utility-margin statistics are
recomputed using the enlarged observation set.

Among the 200 GPQA training queries, 50 queries stop after the
\(2\times2\) stage, while the remaining 150 queries use the
\(3\times3\) budget.
The resulting average number of observations per query--model pair is
therefore
\begin{equation}
\frac{50\times4+150\times9}{200}
=
7.75.
\end{equation}
Compared with fixed \(3\times3\) sampling, which uses nine observations
for every query--model pair, adaptive sampling reduces the number of
candidate-model calls by
\begin{equation}
1-\frac{7.75}{9}
=
13.9\%.
\end{equation}
The recorded completion-token reduction is only \(0.0699\%\), indicating
that the main efficiency gain in this configuration comes from reducing
the number of model invocations rather than the total number of generated
tokens.
\begin{table}[t]
\centering
\small
\setlength{\tabcolsep}{7pt}
\caption{
Observation-budget allocation of adaptive DARS on GPQA.
}
\label{tab:adaptive_sampling}
\resizebox{\linewidth}{!}{
\begin{tabular}{lccc}
\toprule
Strategy
& \(2\times2\) Queries
& \(3\times3\) Queries
& Avg. Obs. \\
\midrule
Fixed \(3\times3\)
& 0
& 200
& 9.00 \\
Adaptive DARS
& 50
& 150
& 7.75 \\
\bottomrule
\end{tabular}
}
\end{table}

\paragraph{Sensitivity to decoding temperature.}
The main experiments use temperature \(T=0.7\) and top-\(p=0.95\)
for stochastic decoding.
Since temperature directly affects the output distribution of an LLM,
we further examine whether the output-side instability identified in
Section~3.3 depends on this particular decoding configuration.
We vary the decoding temperature over
\[
T\in\{0.3,0.5,0.7,0.9\},
\]
while keeping the training data, candidate model pool, query rewriting
procedure, router configuration, and all other decoding parameters
unchanged.

For each temperature, we recompute output-side uncertainty and the winner
flip rate using the same diagnostic protocol as in Section~3.3.
Table~\ref{tab:temperature_sensitivity} reports the complete results.
Both output-side uncertainty and winner instability remain substantial
across all tested temperatures.
Moreover, neither metric changes monotonically with temperature, and the
default setting \(T=0.7\) does not produce the largest output-side
uncertainty.
These results indicate that the instability of single-shot supervision is
not an artifact of a particular decoding temperature.
\begin{table}[t]
\centering
\small
\setlength{\tabcolsep}{8pt}
\caption{
Sensitivity of output-side uncertainty and winner instability to
decoding temperature.
\(T=0.7\) is the default setting used in the main experiments.
}
\label{tab:temperature_sensitivity}
\resizebox{\linewidth}{!}{
\begin{tabular}{ccc}
\toprule
Temperature
& Output-side Uncertainty
& Winner Flip Rate \\
\midrule
0.3 & 0.2162 & 0.8800 \\
0.5 & 0.1993 & 0.7000 \\
0.7 & 0.1939 & 0.8800 \\
0.9 & 0.2143 & 0.9000 \\
\bottomrule
\end{tabular}
}
\end{table}

\paragraph{Sensitivity to the risk coefficient.}
The risk coefficient \(\beta\) controls the strength of the penalty on
performance variability in the DARS utility,
\begin{equation}
U(x,m)
=
\mu_q(x,m)
-
\lambda\mu_c(x,m)
-
\beta\sigma_q(x,m).
\end{equation}
We evaluate
\[
\beta
\in
\{0,\,0.01,\,0.02,\,0.05,\,0.1,\,0.2,\,0.5\}
\]
while keeping all other experimental settings unchanged.
As shown in Figure~\ref{fig:beta_sensitivity}, DARS remains generally
stable over a broad range of risk coefficients.
Among the tested settings, \(\beta=0.2\) provides a strong and stable
choice across the three datasets, and we therefore use it as the default
in the main experiments.

\subsection{Detailed Explanation of the Diagnostic Analysis}
\label{app:diagnostic-analysis}

This section provides additional details on the diagnostic analysis in Section~3.3. The goal of this analysis is to examine whether the standard single-shot supervision protocol provides reliable routing labels. We focus on three aspects: whether the observed outcome of a query-model pair is stable, whether the best model selected for a query is stable, and whether single-shot routing labels agree with labels estimated from repeated observations.

For each query $x_i$ and candidate model $m \in \mathcal{M}$, we collect a set of repeated observations through query rewriting and stochastic decoding. Specifically, let
\[
\mathcal{O}_{i,m}=\{(q^{(n,j)}_{i,m}, c^{(n,j)}_{i,m})\}_{n=1,j=1}^{N,M}
\]
denote the observations of model $m$ on query $x_i$, where $n$ indexes rewritten query variants and $j$ indexes independent decoding runs. Here, $q^{(n,j)}_{i,m}$ is the task-specific quality score and $c^{(n,j)}_{i,m}$ is the corresponding normalized inference cost. We also define the per-observation utility as
\[
u^{(n,j)}_{i,m}=q^{(n,j)}_{i,m}-\lambda c^{(n,j)}_{i,m}.
\]

\paragraph{Outcome instability.}
Outcome instability measures whether repeated observations of the same query-model pair lead to different observed performance scores. For each query-model pair, we check whether the quality scores vary across repeated observations:

{\footnotesize
\[
\mathrm{Instab}_{\mathrm{out}}(i,m)
=
\mathbb{I}
\left[
\max_{n,j} q^{(n,j)}_{i,m}
-
\min_{n,j} q^{(n,j)}_{i,m}
>
\epsilon
\right],
\]
}
where $\epsilon$ is a small tolerance used to avoid numerical artifacts. The overall outcome instability is computed by averaging this indicator over all queries and candidate models:

{\footnotesize
\[
\mathrm{OutcomeInstability}
=
\frac{1}{|\mathcal{D}||\mathcal{M}|}
\sum_{i=1}^{|\mathcal{D}|}
\sum_{m\in\mathcal{M}}
\mathrm{Instab}_{\mathrm{out}}(i,m).
\]
}
A higher value means that the same model can receive different scores on the same underlying query under natural input or output variations.

\paragraph{Winner flip rate.}
Outcome instability at the query-model level does not necessarily imply that the routing label changes. Therefore, we further measure whether the model selected as the best candidate for a query changes across repeated single-shot observations. For each observation index $(n,j)$, we construct a single-shot routing label by selecting the model with the largest observed utility:
\[
w_i^{(n,j)}
=
\arg\max_{m\in\mathcal{M}}
u^{(n,j)}_{i,m}.
\]
The winner flip indicator for query $x_i$ is then defined as

{\footnotesize
\[
\mathrm{Flip}(i)
=
\mathbb{I}
\left[
\left|
\{w_i^{(n,j)}: 1\leq n\leq N, 1\leq j\leq M\}
\right|
> 1
\right].
\]
}
The winner flip rate is the average of this indicator over all queries:
\[
\mathrm{WinnerFlipRate}
=
\frac{1}{|\mathcal{D}|}
\sum_{i=1}^{|\mathcal{D}|}
\mathrm{Flip}(i).
\]
This metric captures whether different single-shot samples can lead to different best-model labels for the same query.

\paragraph{Single-shot versus distribution-aware disagreement.}
We also compare the single-shot routing label with a distribution-aware label estimated from all repeated observations. For each query-model pair, we first compute the expected quality and expected cost:

\[
\mu_q(i,m)=\frac{\sum_{n=1}^{N}\sum_{j=1}^{M}q^{(n,j)}_{i,m}}{NM},
\]
\[\mu_c(i,m)=\frac{\sum_{n=1}^{N}\sum_{j=1}^{M}c^{(n,j)}_{i,m}}{NM}.
\]
The distribution-aware utility used in the diagnostic analysis is
\[
\bar{u}_{i,m}=\mu_q(i,m)-\lambda\mu_c(i,m),
\]
and the corresponding distribution-aware routing label is
\[
w_i^{\mathrm{DA}}
=
\arg\max_{m\in\mathcal{M}}
\bar{u}_{i,m}.
\]
For each single-shot observation $(n,j)$, we compute whether its selected model disagrees with the distribution-aware label:
\[
\mathrm{Disagree}^{(n,j)}(i)
=
\mathbb{I}
\left[
w_i^{(n,j)} \neq w_i^{\mathrm{DA}}
\right].
\]
The final disagreement rate is averaged over all queries and single-shot observations:
\[
\frac{\sum_{i=1}^{|\mathcal{D}|}
\sum_{n=1}^{N}
\sum_{j=1}^{M}
\mathrm{Disagree}^{(n,j)}(i).}{|\mathcal{D}|NM}
\]
This metric measures how often a routing label constructed from one sampled observation differs from the label obtained by aggregating repeated observations.

\paragraph{Router variance under single-shot supervision.}
Finally, we examine whether the instability of single-shot labels propagates to the learned router. We construct multiple single-shot training sets by randomly sampling one observation for each query-model pair. Each sampled training set is used to train the same router architecture under the same data split and model pool. Since the only changing factor is the sampled supervision signal, the variation in test utility reflects the sensitivity of learned routing policies to single-shot supervision noise.

\section{Validity Checks}
\label{app:validity}

DARS relies on two preprocessing components: query rewriting for constructing input-side variations, and automatic scoring for evaluating repeated model outputs. We therefore conduct additional validity checks to examine whether these components introduce obvious artifacts. These checks are intended as lightweight sanity checks rather than formal guarantees of semantic equivalence or scoring correctness.

\subsection{Rewrite Validity Check}
\label{app:rewrite_validity}

The purpose of query rewriting is to introduce controlled input-side variations while preserving the underlying task semantics. To assess the quality of the generated rewrites, we perform an automatic validity check over all rewritten queries used in our experiments. The check covers five aspects: (1) whether all required rewrites are successfully generated, (2) whether rewritten queries duplicate the original query or duplicate each other, (3) whether dataset-specific structural constraints are preserved, (4) whether the rewritten query remains close to the original query in sentence-embedding space, and (5) whether numeric tokens are preserved.

For structural constraints, GPQA rewrites are required to keep the answer choices unchanged; MATH-500 rewrites are required to preserve the mathematical problem structure; and DROP-800 rewrites are required to keep the passage unchanged and rewrite only the question. Sentence similarity is computed using the same sentence-transformer backend as in our query representation pipeline. We additionally flag rewrites with low semantic similarity or numeric-token mismatch for manual inspection. These flags are conservative: a flagged rewrite is not necessarily invalid, since paraphrasing can legitimately change surface forms, symbols, or numeric formatting. The results are summarized in Table~\ref{tab:rewrite_validity}.

\begin{table*}[t]
\centering
\caption{Automatic validity check for query rewrites. Complete denotes the fraction of original queries for which all rewrites are generated. Original dup. and Rewrite dup. denote duplicate rates with the original query and among rewrites, respectively. Structural pass checks dataset-specific constraints. Review flag marks rewrites selected for further inspection due to low similarity or numeric mismatch.}
\label{tab:rewrite_validity}
\small
\resizebox{\textwidth}{!}{
\begin{tabular}{llcccccccccc}
\toprule
Dataset & Split
& Records
& Rewrites
& Complete
& Orig. dup.
& Rew. dup.
& Sim. mean
& Sim. p05
& Num. preserved
& Struct. pass
& Review flag \\
\midrule
GPQA
& train
& 200
& 1000
& 1.0000
& 0.0000
& 0.0000
& 0.8723
& 0.6267
& 0.6474
& 1.0000
& 0.2910 \\
GPQA
& test
& 248
& 1240
& 1.0000
& 0.0000
& 0.0000
& 0.8815
& 0.6656
& 0.6533
& 1.0000
& 0.3016 \\
MATH-500
& train
& 200
& 1000
& 1.0000
& 0.0000
& 0.0000
& 0.9191
& 0.8034
& 0.8564
& 1.0000
& 0.1350 \\
MATH-500
& test
& 300
& 1500
& 1.0000
& 0.0000
& 0.0000
& 0.9248
& 0.8093
& 0.8615
& 1.0000
& 0.1333 \\
DROP-800
& train
& 200
& 1000
& 1.0000
& 0.0000
& 0.0000
& 0.8875
& 0.7449
& 0.9030
& 1.0000
& 0.0400 \\
DROP-800
& test
& 600
& 3000
& 1.0000
& 0.0000
& 0.0000
& 0.8937
& 0.7451
& 0.8893
& 1.0000
& 0.0467 \\
\bottomrule
\end{tabular}
}
\end{table*}

\begin{table*}[t]
\centering
\caption{Validity check for the automatic scoring pipeline. Parse success measures whether a generated response can be parsed into the expected answer format. Finite and In $[0,1]$ check numerical validity. Score type checks whether the dataset-specific scorer returns the expected score type. Rescore agreement measures agreement between stored scores and scores recomputed by the unified scorer.}
\label{tab:scoring_validity}
\small
\resizebox{\textwidth}{!}{
\begin{tabular}{llcccccccccc}
\toprule
Dataset & Split
& Rows
& Parse success
& Finite
& In $[0,1]$
& Score type
& Rescore agreement
& Rescore diff.
& Duplicate keys \\
\midrule
GPQA
& train
& 30000
& 0.9996
& 1.0000
& 1.0000
& 1.0000
& 0.9998
& 7
& 0 \\
GPQA
& test
& 8928
& 0.9999
& 1.0000
& 1.0000
& 1.0000
& 0.9998
& 2
& 0 \\
MATH-500
& train
& 30000
& 0.9992
& 1.0000
& 1.0000
& 1.0000
& 1.0000
& 0
& 0 \\
MATH-500
& test
& 10800
& 0.9984
& 1.0000
& 1.0000
& 1.0000
& 1.0000
& 0
& 0 \\
DROP-800
& train
& 30000
& 0.9998
& 1.0000
& 1.0000
& 1.0000
& 0.9996
& 13
& 0 \\
DROP-800
& test
& 21600
& 0.9999
& 1.0000
& 1.0000
& 1.0000
& 0.9999
& 2
& 0 \\
\bottomrule
\end{tabular}
}
\end{table*}

The results show that all datasets achieve full rewrite completion, no duplicate rewrites are detected, and all rewrites pass the dataset-specific structural checks. Sentence similarity is also high on average across datasets. DROP-800 has the lowest review-flag rate, while GPQA has a higher numeric-mismatch flag rate. This is expected because scientific questions often contain symbols, quantities, abbreviations, and domain-specific expressions whose surface forms may change under paraphrasing. We therefore treat the review flag as a conservative diagnostic signal rather than direct evidence of invalid rewrites.

\subsection{Scoring Validity Check}
\label{app:scoring_validity}

We also verify the automatic scoring pipeline used to evaluate generated responses. For each stored generation, we recompute the score using the unified dataset-specific scoring script and check whether the output can be parsed, whether the score is finite, whether the score lies in the valid range $[0,1]$, whether the expected score type is produced, whether rescoring agrees with the stored score, and whether duplicate generation keys exist. The results are shown in Table~\ref{tab:scoring_validity}.

The scoring pipeline achieves near-perfect parse coverage across all datasets, and all parsed scores are finite and lie within the valid range. We also find no duplicate generation keys. The small number of rescoring differences is caused by scorer-version inconsistencies in intermediate stored files. To avoid propagating such inconsistencies, all reported experimental results are computed after applying the same unified scoring pipeline to the stored generations.

\FloatBarrier

\section{LLM Usage}

Large language models were used only for language polishing, grammar correction, and wording refinement; the research ideas, method design, experiments, result analysis, and conclusions were completed by the authors.

\end{document}